\documentclass[runningheads]{llncs}
\usepackage[T1]{fontenc}
\usepackage{graphicx,verbatim}

\usepackage{amsmath}
\usepackage{amssymb}
\usepackage{multirow}
\usepackage{xcolor}
\usepackage{colortbl}
\usepackage{booktabs}

\newcommand{\subscript}[1]{{\scriptsize #1}}

\definecolor{bestred}{rgb}{0.85, 0.0, 0.0}
\definecolor{secondblue}{rgb}{0.0, 0.2, 0.75}
\definecolor{interpyes}{rgb}{0.0, 0.55, 0.25}
\definecolor{interpno}{rgb}{0.5, 0.5, 0.5}

\usepackage{booktabs,multirow,adjustbox,array}

\usepackage{hyperref}
\hypersetup{
    colorlinks=false, 
    linkbordercolor={1 0 0}, 
    citebordercolor={0 1 0}, 
    urlbordercolor={0 0 1},  
}

\begin{document}
\title{A Neurosymbolic Framework for Interpretable Skeleton-Based Seizure Detection via Concept-Driven Logical Reasoning}
\titlerunning{Interpretable Concept Driven Seizure Detection}
%


\author{Talha Ilyas\inst{1,2} \and
Deval Mehta\inst{2,3}\thanks{now at School of Computing Technologies, RMIT University, Australia} \and
Zongyuan Ge\inst{2,3}}
\authorrunning{T. Ilyas et al.}
\institute{Department of ECSE, Faculty of Engineering, Monash University, Australia \and
AIM for Health Lab, Faculty of IT, Monash University, Australia \and
Department of DSAI, Faculty of IT, Monash University, Australia\\
\email{\{talha.ilyas@monash.edu, deval.mehta1092@gmail.com\}}}
  
\maketitle              
\begin{abstract}
Video-based seizure detection is essential for the management of epilepsy patients, offering a non-invasive complement to electroencephalography. While several deep learning approaches have been developed for video-based seizure detection, none are inherently interpretable, limiting their adoption and translation into clinical practice. We present, to our knowledge, the first exploration of a neurosymbolic framework for video-based seizure detection that directly addresses this gap. Our approach (1) extracts patient-centric skeleton sequences from epilepsy monitoring units via a prompt-guided foundation model, (2) predicts binary spatio-temporal concept activations grounded in clinical motor semiology guidelines, and (3) composes them via differentiable logic into interpretable Boolean rules with auditable contributions. Furthermore, to mitigate false positives arising from the traditional binary formulation (seizure vs.\ non-seizure), we sub-classify non-seizure segments into clinically relevant normal activities, providing the model with fine-grained discriminative supervision. Evaluated on two public seizure video benchmarks, our framework achieves 89.78\% sensitivity with 0.06 false detections per hour on SAHZU and 85.27\%\,/\,0.09 on IEEE, while producing complete three-level interpretability: every prediction decomposes into \textit{which} motor primitives were detected, \textit{how} they were logically composed, and \textit{how much} each rule contributed to the clinical decision. We publicly release all annotations, extracted pose sequences, our data pipeline and code \href{https://github.com/Mr-TalhaIlyas/CDSD/}{Link}.


\end{abstract}
%
%
%

\section{Introduction}
\label{sec:intro}

Epilepsy is a neurological disorder characterized by seizures, affecting approximately 50 million people worldwide~\cite{world_2024}. Optimal diagnosis, treatment, and management of epilepsy patients depends on accurate seizure detection, including precise prediction of onset and offset times and reliable seizure counting. Recently, video-based seizure detection has gained a significant interest, as it can be deployed in both hospital epilepsy monitoring units (EMUs) and out-of-hospital settings, saving time and resources for healthcare professionals and infrastructure~\cite{mehta2023privacy,karacsony2022novel,hou2022automated}.

In clinical practice, neurologists analyze video recordings to identify the observable motor manifestations of seizures, collectively termed \textit{semiology}, following standardized guidelines published by the International League Against Epilepsy (ILAE)~\cite{beniczky2025updated,turek2022seizure,beniczky2022seizure}. This vocabulary defines a taxonomy of motor signs, including \textit{tonic} posturing (sustained muscular contraction), \textit{clonic} jerking (repetitive rhythmic movements), \textit{versive} deviation (forced head or eye turning), and \textit{dystonic} posturing (sustained abnormal limb positions), among others. Crucially, clinicians reason over these observable primitives compositionally: a tonic-clonic seizure, for instance, is identified by the co-occurrence of bilateral limb stiffening followed by rhythmic jerking, not by a single holistic pattern. This structured clinical reasoning process establishes a standardized foundation for explainable decision-making aligned with established clinical guidelines.

A plethora of deep learning approaches have been proposed for video-based seizure detection, including spatiotemporal architectures~\cite{mehta2023privacy,yang2021video,karacsony2022novel,ponnambalam2025privacy} and skeleton-based pipelines~\cite{xu2024vsvig,ilyas2025privacy,hou2022automated,rehman2024enhanced}. Despite strong performance, these methods largely function as black boxes, offering limited interpretability. In clinical epileptology, where decisions rely on understanding seizure semiology and localization, such opacity constrains practical adoption~\cite{jin2014analyzing,saab2024towards}. Models that detect seizures without explicating the underlying motor signs provide limited clinical value. Additionally, most approaches adopt a binary formulation (\textit{seizure} vs.\ \textit{non-seizure}), collapsing heterogeneous non-seizure activity into a single negative class, which increases false positives during movement-intensive non-seizure activity~\cite{saab2024towards}.

In the general computer vision community, Concept Bottleneck Models (CBMs) \cite{koh2020concept,yang2023language} promote interpretability by routing predictions through human-defined concepts. However, primarily designed for image classification, standard CBMs treat concepts independently and do not model their logical composition, crucial for video understanding. Neurosymbolic AI addresses this limitation by integrating neural perception with symbolic reasoning, enabling differentiable Boolean combinations of concepts that mirror human inference~\cite{wang2023learning,vemuri2025logiccbms}. This compositional paradigm aligns naturally with epileptology, where the ILAE semiology vocabulary defines motor primitives over which clinicians reason logically. We therefore propose a neurosymbolic framework that learns Boolean rules over ILAE-grounded motion concepts via a concept bottleneck with differentiable logic layers, enabling clinically interpretable seizure detection. To further reduce false positives inherent in conventional binary (\textit{seizure} vs.\ \textit{non-seizure}) formulations, we introduce fine-grained normal activity subclasses that provide more discriminative supervision.

\noindent\textbf{Our contributions} are: \textbf{(1)} A neurosymbolic framework for seizure detection that learns differentiable Boolean rules over ILAE-grounded spatio-temporal motion concepts. \textbf{(2)} A fine-grained skeleton-based seizure dataset augmenting two public benchmarks~\cite{xu2024vsvig,nt6e-3x56-24} with eight normal activity subclasses via a patient-focused pose extraction pipeline (SAM~3~\cite{carion2025sam} + Sapiens~\cite{khirodkar2024sapiens}) and VLM-assisted annotation, publicly released. \textbf{(3)} Empirical evidence that fine-grained supervision combined with neurosymbolic reasoning reduces false positives on movement-intensive non-seizure segments while preserving interpretability.

\section{Method}
\label{sec:method}

\begin{figure}[!t]
    \centering
    \includegraphics[width=1.0\linewidth]{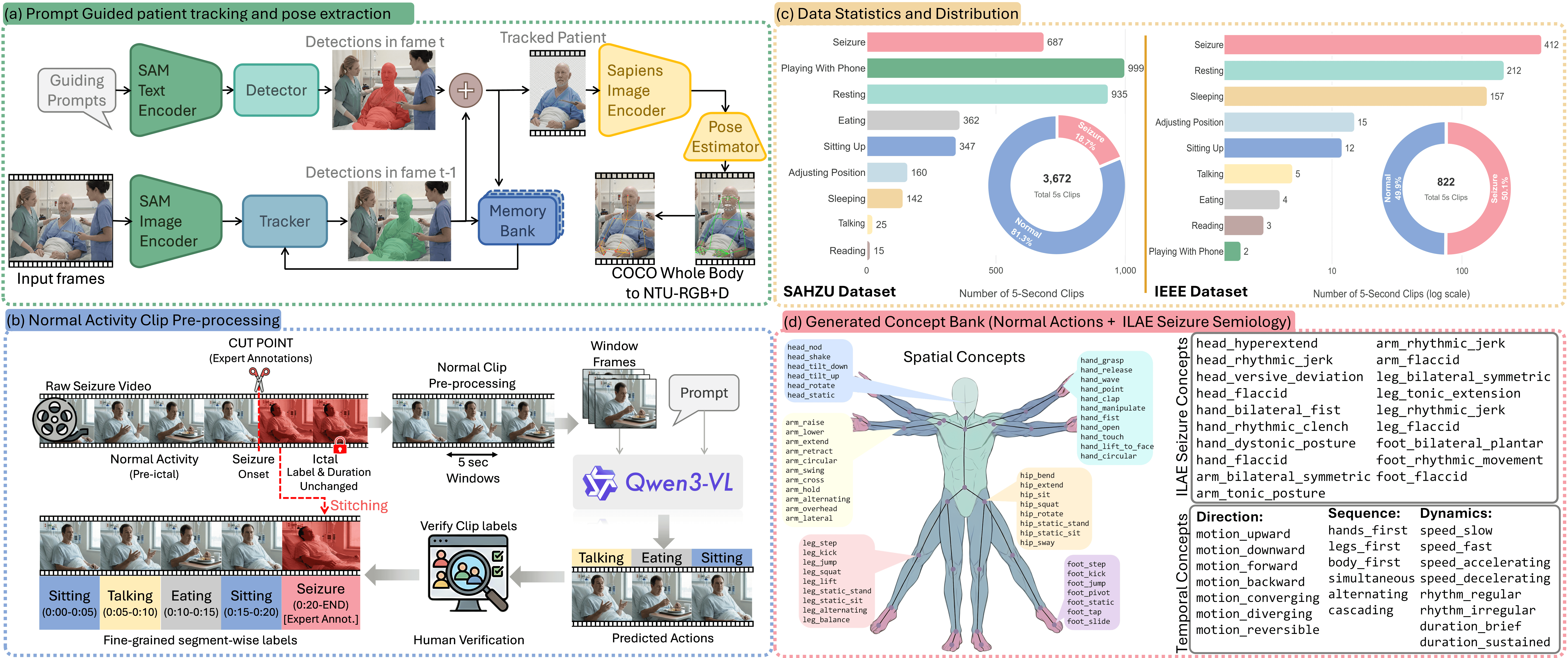}
    \caption{(a) Prompt guided patient tracking and pose extraction, (b) VLM assisted normal activity sub-classificaiton and verification, (c) Data Classes, statistics and distribution, (d) Generated concept bank for NTU-RGB+D 120, SAHZU and IEEE Dataset.}
    \label{fig:data}
\end{figure}

Our framework comprises three stages (Fig.~\ref{fig:data}): (1)~patient-centric skeleton extraction and fine-grained activity labeling from raw EMU video (\S\ref{sec:data_prep}); (2)~construction of a unified concept bank grounded in ILAE motor semiology (\S\ref{sec:concept_bank}); and (3)~neurosymbolic inference via differentiable logic layers (\S\ref{sec:model}).

\subsection{Data Pre-processing and Fine-grained Label Generation}
\label{sec:data_prep}

We evaluate on two public seizure video datasets: the SAHZU dataset~\cite{xu2024vsvig} and the IEEE Seizure Video dataset~\cite{nt6e-3x56-24}. Both provide expert-labeled seizure onset/offset annotations but only binary (\textit{seizure}/\textit{non-seizure}) labels.

\noindent\textbf{Patient Tracking and Pose Extraction}
\label{sec:pose_extraction}
EMU recordings differ substantially from standard human action recognition (HAR) settings: patients are typically supine, partially occluded by bedding, and share the frame with medical staff. To bridge this gap, we convert EMU videos into a standardized HAR representation using a two-stage foundation model pipeline (Fig.~\ref{fig:data}(a)). First, SAM~3~\cite{carion2025sam} segments and tracks the patient across frames using text-prompt initialization and streaming memory to maintain identity under transient occlusions. The segmented regions are then processed by Sapiens-2B~\cite{khirodkar2024sapiens} to extract 133 COCO-WholeBody keypoints~\cite{jin2020whole}, which we map to the 25-joint NTU~RGB+D format~\cite{liu2019ntu} via $\Psi{:}\mathbb{R}^{133 \times 2} \rightarrow \mathbb{R}^{25 \times 2}$. This alignment enables direct use of pretrained GCN backbones, crucial given the limited scale of seizure datasets.

\noindent\textbf{Fine-Grained Normal Activity Annotation}
\label{sec:fine_grained}
Conventional binary seizure/non-seizure annotation collapses heterogeneous normal activities into a single class, inflating false positives on movement-intensive segments (\textit{e.g.}, patient adjusting position) that share superficial kinematic similarity with seizure activity~\cite{saab2024towards}. To increase class specificity, we thus sub-classify non-seizure segments into fine-grained normal activities while preserving expert-labeled seizure boundaries untouched. Videos are clipped at the annotated seizure onset; only non-seizure segments are divided into 5-second clips and classified by Qwen3-VL~\cite{bai2025qwen3} using structured prompts with clinical context. Predictions are then verified by human annotators and stitched back with the original seizure labels, yielding 8 non-seizure activity categories (Fig.~\ref{fig:data}(c)), 4 of which overlap with NTU~RGB+D~120~\cite{liu2019ntu} (Fig.~\ref{fig:data}(d)), facilitating knowledge transfer from large-scale HAR pretraining.

\subsection{Seizure and Normal Activity Concept Bank Generation}
\label{sec:concept_bank}

Seizure detection requires a concept vocabulary spanning two distinct behavioral regimes: normal daily activities and pathological motor semiology. We construct a unified concept bank $\mathcal{C} = \mathcal{C}_\text{s} \cup \mathcal{C}_\text{t}$ of spatial and temporal concepts in two phases, following the concept bottleneck paradigm~\cite{koh2020concept}.

\noindent\textbf{Phase~1: Normal activity concepts.} Understanding normal activity primitives is essential for distinguishing seizure-specific motion patterns. To establish a comprehensive concept bank, we leverage the large-scale NTU RGB+D 120 dataset, which contains 120 diverse human action classes. Following~\cite{li2022hake}, each action is decomposed into six body-part groups $\mathcal{P}{=}{\texttt{head}, \texttt{hand}, \texttt{arm}, \texttt{hip}, \texttt{leg}, \texttt{foot}}$, and GPT-4o is prompted to generate part-wise motion descriptions. These descriptions are abstracted into canonical $\texttt{verb-direction-modifier}$ patterns, embedded using a sentence encoder~\cite{reimers-2019-sentence-bert}, and clustered via $k$-means to form spatial concepts $\mathcal{C}\text{s}^{\text{norm}}$, and temporal concepts $\mathcal{C}\text{t}^{\text{norm}}$ capturing motion dynamics (Fig.~\ref{fig:data}(d)). This process yields 74 concepts (53 spatial, 21 temporal).

\noindent\textbf{Phase~2: Seizure-specific concepts from ILAE semiology.} Seizure movements follow standardized clinical definitions that LLM-generated descriptions alone cannot capture. We treat seizure as a single holistic category and derive 19 spatial concepts from the ILAE glossary~\cite{beniczky2025updated,turek2022seizure,beniczky2022seizure} using the same $\mathcal{P}$ partitioning, covering shared motor signatures across tonic, clonic, myoclonic, atonic, dystonic, and versive semiology (\textit{e.g.}, \texttt{arm\_tonic\_posture}, \texttt{hand\_bilateral\_fist}, \texttt{head\_versive\_deviation}). The 21 temporal concepts from Phase~1 are shared across both regimes, as they already encode dynamics relevant to seizure discrimination Fig.\ref{fig:data}(d). This yields a unified bank of $|\mathcal{C}|{=}93$ concepts (53 normal + 19 seizure spatial, 21 shared temporal).

\noindent\textbf{Association matrix.} We construct $\mathbf{M} \in \{0,1\}^{|\mathcal{A}| \times |\mathcal{C}|}$ where $\mathbf{M}_{a,c}{=}1$ indicates concept $c$ is active for action $a$. Associations are generated via an LLM~\cite{grattafiori2024llama} and refined pairwise to ensure discriminability, \textit{e.g.}, \texttt{adjusting\_position} and \texttt{seizure} share \texttt{arm\_extend} but differ on seizure-specific concepts such as \texttt{arm\_tonic\_posture}.
\begin{figure}[!t]
    \centering
    \includegraphics[width=1.0\linewidth]{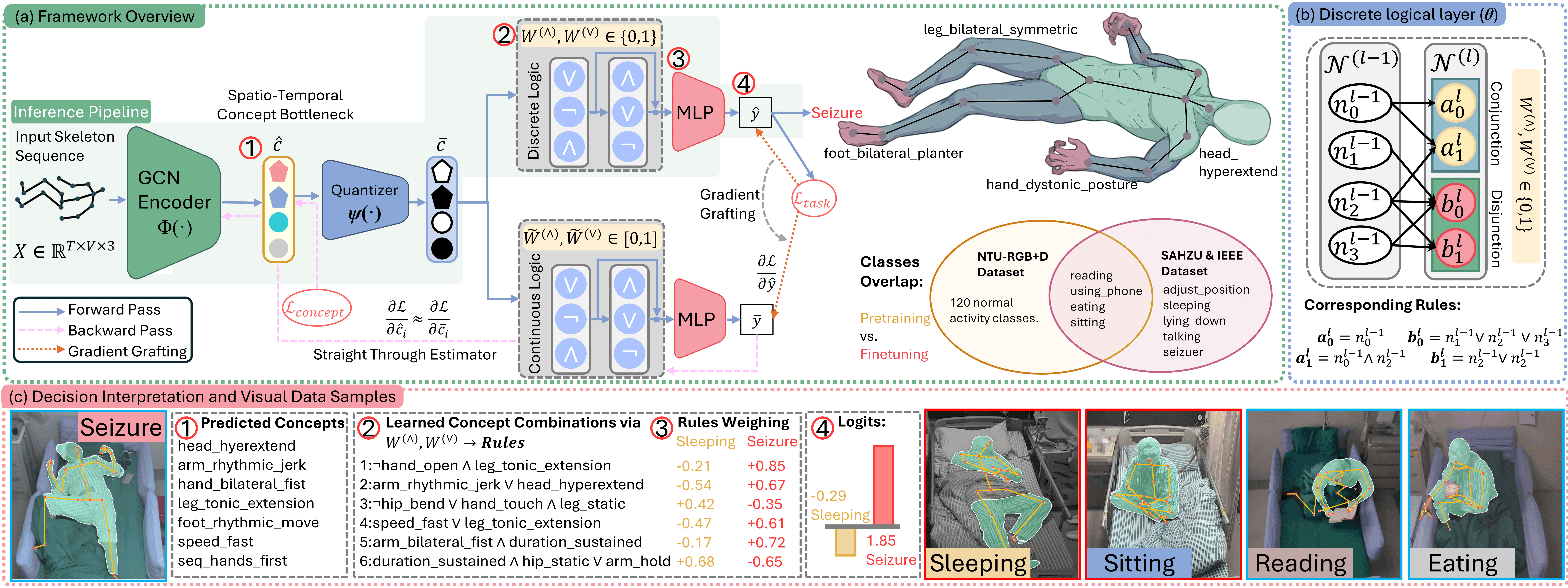}
    \caption{(a) Model Framework overview along with body part specific concept of seizure dataset, (b) Discrete logic layer example along with it's corresponding rules, (c) Visual samples of normal and seizure activities along with rule interpretation guide.}
    \label{fig:model}
\end{figure}

\subsection{Neurosymbolic Seizure Detection Framework}
\label{sec:model}
Our framework (Fig.~\ref{fig:model}(a)) processes each skeleton sequence through four stages: a \textit{spatio-temporal encoder} $\Phi$ extracts motion features, a \textit{concept bottleneck} $\psi$ maps them to binary concept activations, \textit{differentiable logic layers} $\theta$ compose concepts into Boolean rules, and a \textit{rule aggregator} produces class logits as a weighted sum of triggered rules.

\textbf{Skeleton Encoding and Concept Bottleneck:}
\label{sec:encoder_concept}
Given an input sequence $\mathbf{X} \in \mathbb{R}^{T \times V \times 3}$ ($T$ frames, $V{=}25$ joints), the encoder $\Phi$ models the skeleton as a spatio-temporal graph via a GCN backbone (\textit{e.g.}, Hyper-GCN~\cite{zhou2025adaptive}), producing features $\mathbf{F} \in \mathbb{R}^{T \times V \times D}$. These are globally averaged into $\bar{\mathbf{F}} \in \mathbb{R}^{D}$ and projected to concept activations:
\begin{equation}
    \hat{\mathbf{c}} = \sigma(\mathbf{W}_c \, \bar{\mathbf{F}} + \mathbf{b}_c) \in [0,1]^{|\mathcal{C}|},
    \label{eq:concept_pred}
\end{equation}
where each $\hat{c}_i$ is the predicted probability that concept $c_i$ is active. The bottleneck is supervised via binary cross-entropy against $\mathbf{c}^* = \mathbf{M}[a,:] \in \{0,1\}^{|\mathcal{C}|}$ from the action label $a$, forcing the encoder to ground representations in interpretable motion primitives before downstream reasoning. To mitigate concept leakage~\cite{gao2025learning,vemuri2025logiccbms}, we binarize activations via $\bar{c}_i = \mathbb{1}[\hat{c}_i > 0.5]$ and augment the predicate space with negations: $\mathbf{p} = [\bar{\mathbf{c}};\, \neg\bar{\mathbf{c}}] \in \{0,1\}^{2|\mathcal{C}|}$, enabling rules over both presence and absence of motion concepts.

\textbf{Differentiable Logic Layers:}
\label{sec:logic_layers}
The logic layers $\theta$ compose binary predicates into rules via stacked conjunction-disjunction operations (Fig.~\ref{fig:model}(b)). Let $\mathbf{n}^{(0)} = \mathbf{p}$ denote the input predicate vector. Each layer $l$ produces two sets of outputs: conjunction nodes $a_i^{(l)}$ that represent candidate AND-rules over predicates from the previous layer, and disjunction nodes $b_i^{(l)}$ that represent candidate OR-rules: 

\begin{equation}
a_i^{(l)} = \bigwedge_{j:\, W_{ij}^{(l, \land)} = 1} n_j^{(l-1)}, \quad
b_i^{(l)} = \bigvee_{j:\, W_{ij}^{(l, \lor)} = 1} n_j^{(l-1)}, \label{eq:discrete_logic}
\end{equation}

\noindent where $\mathbf{W}^{(l,\land)}, \mathbf{W}^{(l,\lor)} \in \{0,1\}^{N_l \times N_{l-1}}$ are learnable binary adjacency matrices that select which predicates participate in each rule. The concatenated output $\mathbf{n}^{(l)} = [\mathbf{a}^{(l)};\, \mathbf{b}^{(l)}]$ feeds subsequent layers, progressively building rules of increasing complexity in disjunctive normal form. After $L$ layers, the final $\mathbf{r} \in \{0,1\}^R$ contains activations of $R$ learned rules, \textit{e.g.}, $(\texttt{head\_hyperextend} \land \texttt{dynamics\_rhythm\_irregular})$ for seizure patterns.

\textbf{Continuous relaxation and training:}
Since discrete Boolean operations yield zero gradients, we maintain continuous counterparts $\tilde{\mathbf{W}} \in [0,1]$ with differentiable logical activations~\cite{wang2023learning}:
\begin{equation}
    \tilde{a}_i^{(l)} = \mathcal{P}\!\Bigl(\prod\nolimits_{j} \bigl(1 - \tilde{W}_{ij}^{(l,\land)}(1 - \tilde{n}_j^{(l-1)})\bigr)\Bigr), \;\;
    \tilde{b}_i^{(l)} = 1 - \mathcal{P}\!\Bigl(\prod\nolimits_{j} \bigl(1 - \tilde{W}_{ij}^{(l,\lor)} \tilde{n}_j^{(l-1)}\bigr)\Bigr),
    \label{eq:soft_logic}
\end{equation}
where $\mathcal{P}(x) = 1/(1{-}\log x)$ prevents gradient vanishing; these reduce to standard AND/OR when inputs and weights are binary. The forward pass uses discrete weights $\mathbf{W} = \mathbb{1}[\tilde{\mathbf{W}} > 0.5]$ for interpretable inference, while backpropagation \textit{grafts} gradients onto the continuous path to update $\tilde{\mathbf{W}}$. A Straight-Through Estimator~\cite{li2022hake} bridges the binarizer, enabling end-to-end gradient flow from the task loss through the logic layers to $\Phi$. The total objective is $\mathcal{L} = \mathcal{L}_\text{task} + \mathcal{L}_\text{concept} + \lambda \|\tilde{\mathbf{W}}\|_2$, where the sparsity term $\lambda$ encourages parsimonious, interpretable rules.


\textbf{Rule Aggregation and Decision Interpretation:}
\label{sec:interpretation}
A linear layer maps rule activations to class logits $\hat{\mathbf{y}} = \mathbf{V}\mathbf{r} + \mathbf{b}$, where each weight $V_{a,k}$ quantifies how rule $r_k$ supports or opposes class $a$, yielding an additive decomposition $\hat{y}_a = \sum_k V_{a,k} \cdot r_k + b_a$. This produces a three-level audit trail (Fig.~\ref{fig:model}(c)): (1)~\textit{concept explanations}, which ILAE-aligned motor primitives are detected, (2)~\textit{rule explanations}, how concepts are composed into Boolean rules by tracing adjacency matrices, and (3)~\textit{rule contributions}, how much each rule influenced the prediction via weights $V_{a,k}$.

\section{Experimental Setup}
\label{sec:experiments}



\noindent\textbf{Datasets:} We evaluate on two public seizure video datasets. \textbf{SAHZU}~\cite{xu2024vsvig} contains EMU recordings of 14 patients with 33 focal or generalized tonic-clonic seizures (~3.3\,h total). \textbf{IEEE Seizure Video}~\cite{nt6e-3x56-24} comprises recordings from 50 patients with 403 seizure segments (334 tonic-clonic, 69 absence) and an equal number of non-seizure segments. We use the official \cite{xu2024vsvig} 70/30 split for SAHZU and report 5-fold cross-validation for IEEE.

\noindent\textbf{Preprocessing.}
We follow the standard skeleton-based HAR preprocessing protocol from CTR-GCN~\cite{chen2021channel}: random rotation, spatial flip, and temporal cropping with uniform 64-frame sampling. Since our pose extraction pipeline (\S\ref{sec:pose_extraction}) outputs skeletons in the NTU-RGB+D 25-joint format, the same preprocessing applies directly to our seizure data without modification, ensuring reproducibility and compatibility with established baselines.


\noindent\textbf{Pretraining on NTU-RGB+D~120:} The limited scale of seizure datasets makes training from scratch impractical. We first pretrain the encoder $\Phi$ and concept bottleneck $\psi$ on NTU-RGB+D 120~\cite{liu2019ntu} using our 74 normal-activity concepts, achieving 86.7\% on the cross-subject benchmark, on par with the state of the art~\cite{zhou2025adaptive,ilyas2026neurosymbolic}. This confirms the concept vocabulary captures sufficient discriminative information before extending it with seizure-specific concepts for clinical transfer.

\noindent\textbf{Fine-tuning and logic layer training:}
Starting from the pretrained encoder, we fine-tune the full framework on the seizure datasets for 400 epochs using AdamW with cosine learning rate decay on a single NVIDIA A6000 GPU with batch size 32. The concept bank is extended with the 19 seizure-specific concepts (\S\ref{sec:concept_bank}), and the concept predictor head is expanded accordingly. Logic layers are activated after 15 warm-up epochs to ensure they receive converged concept predictions rather than random activations. The learning rate is set to $10^{-4}$ for the entire framework, with $\ell_1$ regularization ($\lambda{=}10^{-5}$) inducing rule sparsity.

\section{Results}
\label{sec:results}

\textbf{Comparison with state-of-the-art methods:} Table~\ref{tab:benchmark} compares our framework against 18 baselines. Optical-flow and RGB-based methods suffer elevated false detection rates (${\geq}0.72$/h) due to EMU noise, while multimodal approaches still lag behind skeleton-only models. Our framework surpasses the strongest skeleton baseline HyperGCN~\cite{zhou2025adaptive} by \textbf{+7.37} sensitivity and \textbf{+5.45} F1 on SAHZU with 68\% FDR reduction (0.19$\to$\textbf{0.06}), and by \textbf{+5.74}\,/\,\textbf{+4.93} on IEEE with 63\% FDR reduction. Against VSViG~\cite{xu2024vsvig}, we gain +8.24\% sensitivity with a $2.3\times$ smaller model. Ours is the \textit{only} method providing full interpretability.

\begin{table*}[!t]
\centering
\caption{Comparison with SOTA methods on SAHZU and IEEE benchmarks. \textcolor{bestred}{Red} = best; \textcolor{secondblue}{\underline{blue}} = second best; Interp.=interpretability; and OF=Optical Flow.}
\label{tab:benchmark}
\resizebox{0.9\textwidth}{!}{%
\begin{tabular}{llccccccccccc}
\toprule
\multirow{2}{*}{\textbf{Method}} & \multirow{2}{*}{\textbf{Modality}} & \multicolumn{4}{c}{\textbf{SAHZU}} & \multicolumn{4}{c}{\textbf{IEEE}} & \multirow{2}{*}{\textbf{Interp.}} & \textbf{Param.} \\ \cmidrule(lr){3-6} \cmidrule(lr){7-10}
 & & \textbf{Sens.\%} & \textbf{Spec.\%} & \textbf{F1\%} & \textbf{FDR/h} & \textbf{Sens.\%} & \textbf{Spec.\%} & \textbf{F1\%} & \textbf{FDR/h} &  & \textbf{(M)} \\ \midrule
%
%
I3D+LSTM~\cite{karacsony2022novel}
    & OF & 71.23 & 68.45 & 69.81 & 0.87
    & 67.84 & 65.12 & 66.43 & 1.04
    & \textcolor{interpno}{NO} & 28.0 \\
CNN+LSTM~\cite{yang2021video}
    & OF & 68.36 & 65.92 & 67.08 & 1.13
    & 64.51 & 62.87 & 63.62 & 1.31
    & \textcolor{interpno}{NO} & 17.1 \\
R3D+LSTM~\cite{perez2021transfer}
    & OF & 72.68 & 69.37 & 70.94 & 0.79
    & 69.15 & 66.48 & 67.74 & 0.95
    & \textcolor{interpno}{NO} & 31.9 \\
R3D+ViT~\cite{mehta2023privacy}
    & OF & 74.15 & 71.83 & 72.92 & 0.72
    & 70.87 & 68.24 & 69.48 & 0.86
    & \textcolor{interpno}{NO} & 32.6 \\ \midrule
%
%
JOLOGCN~\cite{xu2024vsvig}
    & OF+Skel. & 75.61 & 73.28 & 74.38 & 0.52
    & 72.34 & 70.15 & 71.17 & 0.64
    & \textcolor{interpno}{NO} & 6.9 \\
RGBPoseConv3D~\cite{xu2024vsvig}
    & RGB+Skel. & 76.93 & 74.67 & 75.74 & 0.44
    & 73.48 & 71.52 & 72.43 & 0.56
    & \textcolor{interpno}{NO} & 3.2 \\
AGCN+TCN~\cite{hou2022automated}
    & RGB+Skel. & 74.82 & 72.16 & 73.41 & 0.58
    & 71.37 & 69.24 & 70.23 & 0.71
    & \textcolor{interpno}{NO} & 37.1 \\
RAMI~\cite{ilyas2025privacy}
    & OF+Skel. & 78.54 & 76.83 & 79.77 & 0.61
    & 75.16 & 73.42 & 76.28 & 0.74
    & \textcolor{interpno}{NO} & 34.46 \\
VSViG~\cite{xu2024vsvig}
    & RGB+Patch & 81.54 & 79.72 & 80.77 & \textcolor{secondblue}{\underline{0.09}}
    & 78.62 & 77.87 & \textcolor{secondblue}{\underline{79.53}} & \textcolor{secondblue}{\underline{0.12}}
    & \textcolor{interpno}{NO} & 5.4 \\ \midrule
%
%
STGCN~\cite{liu20253d}
    & Skeleton & 76.47 & 74.31 & 75.32 & 0.38
    & 73.82 & 71.65 & 72.67 & 0.46
    & \textcolor{interpno}{NO} & 3.1 \\
DG-STGCN~\cite{liu20253d}
    & Skeleton & 78.24 & 76.18 & 77.15 & 0.31
    & 75.63 & 73.71 & 74.59 & 0.38
    & \textcolor{interpno}{NO} & 1.4 \\
MSG3D~\cite{liu20253d}
    & Skeleton & 79.86 & 77.54 & 78.63 & 0.27
    & 76.91 & 74.82 & 75.79 & 0.34
    & \textcolor{interpno}{NO} & 2.7 \\
AAGCN~\cite{liu20253d}
    & Skeleton & 78.53 & 76.42 & 77.41 & 0.33
    & 75.28 & 73.36 & 74.24 & 0.41
    & \textcolor{interpno}{NO} & 3.7 \\
CTRGCN~\cite{chen2021channel}
    & Skeleton & 81.35 & 79.84 & 80.53 & 0.22
    & 78.47 & 76.92 & 77.62 & 0.28
    & \textcolor{interpno}{NO} & 1.4 \\
HDGCN~\cite{xu2024vsvig}
    & Skeleton & 80.72 & 78.63 & 79.61 & 0.24
    & 77.84 & 75.91 & 76.81 & 0.30
    & \textcolor{interpno}{NO} & 1.7 \\
InfoGCN~\cite{chi2022infogcn}
    & Skeleton & 80.18 & 78.27 & 79.16 & 0.26
    & 77.23 & 75.48 & 76.28 & 0.32
    & \textcolor{interpno}{NO} & 1.57 \\
HyperGCN~\cite{zhou2025adaptive}
    & Skeleton & \textcolor{secondblue}{\underline{82.41}} & \textcolor{secondblue}{\underline{80.67}} & \textcolor{secondblue}{\underline{81.48}} & 0.19
    & \textcolor{secondblue}{\underline{79.53}} & \textcolor{secondblue}{\underline{77.94}} & 78.66 & 0.24
    & \textcolor{interpno}{NO} & 2.4 \\ \midrule
%
%
\begin{tabular}[c]{@{}l@{}}\textbf{Proposed}\\ \subscript{HyperGCN+Logic Layers+Multi-label}\end{tabular}
    & Skeleton & \textcolor{bestred}{89.78} & \textcolor{bestred}{87.64} & \textcolor{bestred}{86.93} & \textcolor{bestred}{0.06}
    & \textcolor{bestred}{85.27} & \textcolor{bestred}{83.28} & \textcolor{bestred}{83.59} & \textcolor{bestred}{0.09}
    & \textcolor{interpyes}{\textbf{YES}} & 2.3 \\ \bottomrule
\end{tabular}%
}
\end{table*}

\textbf{Ablation studies:} Table~\ref{tab:ablation} isolates each design choice. \textit{(a)} NTU-120 pretraining is essential (without it, sensitivity drops 11.16 points, FDR/h degrades $22\times$). The multi-label formulation adds +6.47 sensitivity with $3.5\times$ FDR reduction, validating class specificity~\cite{saab2024towards} in the video modality. \textit{(b)} Our SAM~2 pipeline achieves +9.51 sensitivity over VSViG~\cite{xu2024vsvig} pose extraction with zero manual annotation. \textit{(c)} Temporal concepts boost sensitivity by 9.17 points, capturing rhythmicity and contraction dynamics missed by spatial concepts alone. \textit{(d)} Our framework detects seizures 3.85\,s before clinical onset on SAHZU (1.47\,s on IEEE); VSViG achieves earlier detection (6.8\,s\,/\,3.51\,s) via accumulative probability, but without interpretable explanations.

\begin{table}[!t]
\caption{Ablation studies on the SAHZU dataset.}
\label{tab:ablation}
\resizebox{\textwidth}{!}{%
\begin{tabular}{lcccllcccllccc}
\multicolumn{4}{l}{\textbf{(a) Model Configuration}} & \textbf{} & \multicolumn{4}{l}{\textbf{(b) Patient Tracking Pipeline}} & \textbf{} & \multicolumn{4}{l}{\textbf{(c) Concept Configuration}} \\ \cline{1-4} \cline{6-9} \cline{11-14} 
\textbf{Variant} & \textbf{Sens. \%} & \textbf{F1 \%} & \textbf{FDR/h} & \textbf{} & \textbf{Variant} & \textbf{Sens. \%} & \textbf{F1 \%} & \textbf{FDR/h} & \textbf{} & \textbf{Variant} & \textbf{Sens. \%} & \textbf{F1 \%} & \textbf{FDR/h} \\ \cline{1-4} \cline{6-9} \cline{11-14} 
w/o NTU pretraining & 78.62 & 76.34 & 1.34 &  & Ilyas~et~al.~\cite{ilyas2025privacy} & 85.84 & 83.97 & 0.65 &  & Spatial Only & 80.61 & 81.43 & 0.25 \\
w NTU pretraining & 89.78 & 87.64 & 0.06 &  & Xu~et~al.~\cite{xu2024vsvig} & 80.27 & 81.48 & 0.11 &  & Spatial + Temporal & 89.78 & 87.64 & 0.06 \\ \cline{1-4} \cline{11-14} 
Binary model & 83.31 & 81.26 & 0.21 &  & Proposed & 89.78 & 87.64 & 0.06 &  & \textbf{(d) Latency (sec)} & \textbf{VSViG} & \multicolumn{2}{c}{\textbf{IEEE}} \\ \cline{6-9} \cline{11-14} 
Multilabel model & 89.78 & 87.64 & 0.06 &  &  & \multicolumn{1}{l}{} & \multicolumn{1}{l}{} & \multicolumn{1}{l}{} &  & Xu~et~al.~\cite{xu2024vsvig} & -6.8 & \multicolumn{2}{c}{-3.51} \\ \cline{1-4}
 & \multicolumn{1}{l}{} & \multicolumn{1}{l}{} & \multicolumn{1}{l}{} &  &  & \multicolumn{1}{l}{} & \multicolumn{1}{l}{} & \multicolumn{1}{l}{} &  & Proposed & -3.85 & \multicolumn{2}{c}{-1.47} \\ \cline{11-14} 
\end{tabular}
}
\end{table}

\begin{figure}[!h]
    \centering
    \includegraphics[width=1.0\linewidth]{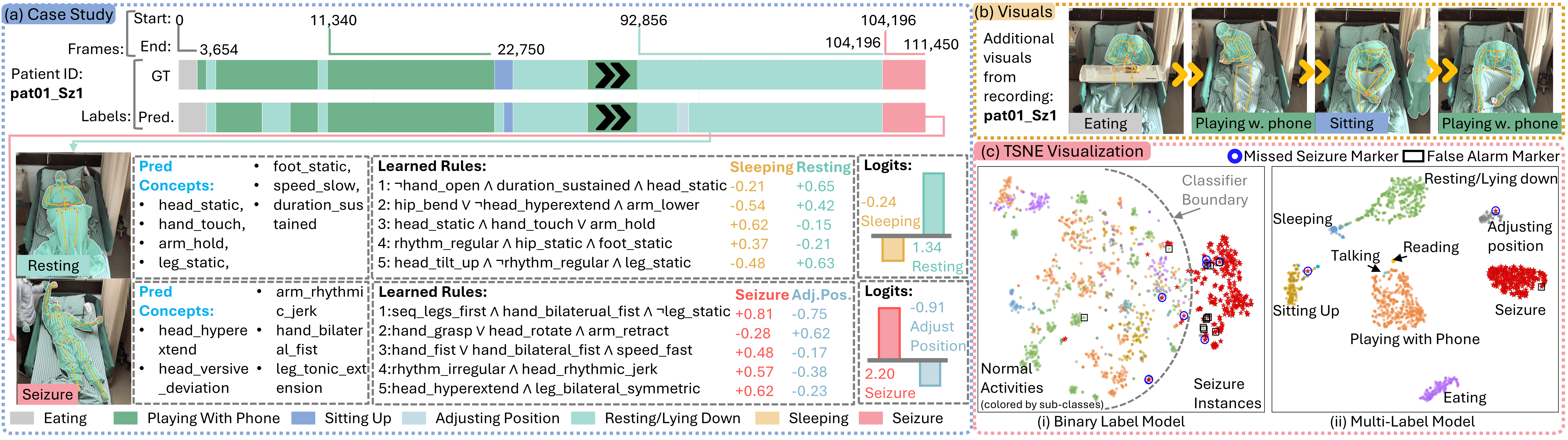}
    \caption{Qualitative analysis on a SAHZU recording. (a)~Ground-truth and predicted timelines with rule contributions for
    representative resting and seizure clips. (b)~Additional frames from an earlier segment. (c)~t-SNE of rule activations: multi-label vs.\ binary formulation.}
    \label{fig:case_study}
\end{figure}

\textbf{Qualitative analysis:} Fig.~\ref{fig:case_study} visualizes our framework on a continuous SAHZU recording. The predicted timeline closely matches the ground truth, with extracted seizure rules aligning with ILAE tonic semiology and resting rules appropriately suppressing other class activations. The t-SNE visualization (Fig.~\ref{fig:case_study}(c)) confirms that multi-label training produces tighter clusters, explaining the $3.5\times$ FDR reduction.

\section{Conclusion}
\label{sec:conclusion}
We introduced the first neurosymbolic framework for video-based seizure detection, grounding every prediction in ILAE motor semiology through a concept bottleneck with differentiable logic layers that produce fully auditable Boolean rules. Our fine-grained formulation for normal activity reduces false detections by 3.5$\times$ over binary classification, while achieving state-of-the-art sensitivity on two public benchmarks. Future work includes fine-grained seizure sub-classification across motor semiology types, expanding seizure-specific temporal concepts beyond the shared vocabulary, and validation on different cohorts.

\subsubsection*{Disclosure of Interests.} The authors declare no competing interests.
\subsubsection*{Acknowledgment.} This work was supported in part by an NVIDIA Academic Grant which provided the cloud access of $8 \times A100$ GPUs from the NVIDIA Corporation.
%
%
%

\bibliographystyle{splncs04}
\bibliography{mybibliography}

\end{document}